\pdfoutput=1
\PassOptionsToPackage{prologue,dvipsnames}{xcolor}
\documentclass[11pt]{article}

\usepackage[]{acl}

\usepackage{times}
\usepackage{latexsym}
\usepackage{hyperref}
\usepackage{url}
\usepackage{graphicx,subcaption}
\usepackage{xspace}
\usepackage{wrapfig}
\usepackage{tikz}
\usepackage{booktabs}
\usepackage{amsmath,amsfonts,bm}
\usepackage{mathtools}
\usepackage{annotate-equations}
\usepackage[dvipsnames]{xcolor}

\usepackage[T1]{fontenc}

\usepackage[utf8]{inputenc}

\usepackage{microtype}

\title{Towards Understanding the Fragility of Multilingual LLMs \\ against Fine-Tuning Attacks}

\author{
    Samuele Poppi$^{2, 3}$\thanks{Work done during internship at Meta.}\quad
    Zheng-Xin Yong$^{4}$\footnotemark[1]\quad
    Yifei He$^{5}$ 
    \\
    {\bf Bobbie Chern}$^{1}$\quad 
    {\bf Han Zhao}$^{5}$\quad
    {\bf Aobo Yang}\thanks{Equal advising.}$^{1}$\quad 
    {\bf Jianfeng Chi}\footnotemark[2]$^{1}$
    \\
    $^{1}$Meta
    \quad
    $^{2}$University of Pisa
    \quad
    $^{3}$University of Modena and Reggio Emilia \\
    \quad
    $^{4}$Brown University
    \quad
    $^{5}$University of Illinois Urbana-Champaign\\
    {\tt samuele.poppi@unimore.it}
    {\tt zheng\_xin\_yong@brown.edu}\\
    {\tt \{yifeihe3, hanzhao\}@illinois.edu} \\
    {\tt \{bgchern, aoboyang, jianfengchi\}@meta.com}
}

\usepackage{graphicx}
\usepackage{subcaption}
\usepackage{mathrsfs}

\def \ie {\emph{i.e.}}
\def \eg {\emph{e.g.}}

\def \wrt {\emph{w.r.t.}}

\newcommand{\model}[1]{\bm{\theta}_{\text{{#1}}}}

\newcommand{\binarymask}[1]{\bm{\gamma}_{\text{{#1}}}}

\newcommand{\localizationfunction}[0]{\mathrm{loc}}
\newcommand{\lpool}[0]{L_{\text{pool}}}

\newcommand{\modelnotext}[1]{\bm{\theta}_{{#1}}}

\newcommand{\binarymasknotext}[1]{\bm{\gamma}_{{#1}}}
\newcommand{\binarymaskwlang}[2]{\binarymasknotext{\text{#1}_{\text{#2}}}}

\newcommand{\ours}[0]{SIL\xspace}

\usepackage{cleveref}
\Crefname{algorithm}{Algorithm}{Algorithms}
\Crefname{appendix}{Appendix}{Appendices}
\Crefname{figure}{Figure}{Figures}
\Crefname{section}{Section}{Sections}
\Crefname{subsection}{Section}{Sections}
\Crefname{subsubsection}{Section}{Sections}
\Crefname{table}{Table}{Tables}

\begin{document}
\maketitle
\begin{abstract}
Recent advancements in Large Language Models (LLMs) have sparked widespread concerns about their safety. Recent work demonstrates that safety alignment of LLMs can be easily removed by fine-tuning with a few adversarially chosen instruction-following examples, \ie, \textit{fine-tuning attacks}. 
We take a further step to understand fine-tuning attacks in multilingual LLMs. We first discover \textit{cross-lingual generalization} of fine-tuning attacks: using a few adversarially chosen instruction-following examples in \textit{one} language, multilingual LLMs can also be easily compromised (\eg, multilingual LLMs fail to refuse harmful prompts in other languages). Motivated by this finding, we hypothesize that safety-related information is language-agnostic and propose a new method termed Safety Information Localization (SIL) to identify the safety-related information in the model parameter space. Through SIL, we validate this hypothesis and find that \textit{only changing 20\% of weight parameters in fine-tuning attacks can break safety alignment across all languages}. Furthermore, we provide evidence to the \textit{alternative pathways} hypothesis for why freezing safety-related parameters does not prevent fine-tuning attacks, and we demonstrate that our attack vector can still jailbreak LLMs adapted to new languages.

\end{abstract}

\section{Introduction}
Large language models (LLMs) have revolutionized the field of artificial intelligence, but their widespread global adoption has also raised concerns about their safety. Despite their numerous benefits, LLMs can produce inaccurate, misleading, or even harmful outputs~\citep{taxonomy2022, llmhallucination2023}. The safety alignment~\citep{ouyang2022training, wei2022finetuned, rafailov2023direct} of LLMs aims to address safety issues by aligning LLMs to produce outputs that are safe, trustworthy and aligned with human values. However, recent studies have demonstrated that the safety-aligned LLMs are not adversarially robust~\citep{zou2023universal, ghanim2024jailbreaking, carlini2024aligned}.
In a seminal work,~\citet{qi2023fine} proposed a fine-tuning attack showing the safety alignment of LLMs can be compromised by fine-tuning only a few steps on a few adversarially designed training examples, either for closed/open-source models~\citep{touvron2023llama, achiam2023gpt}. The fine-tuning attack poses a significant threat to large language models (LLMs) and has led to several follow-up studies~\citep{wei2024assessing, peng2024navigating} aimed at understanding its properties. However, it remains unclear how effective fine-tuning attacks are in multilingual LLMs~\citep{meta2024llama3, yang2024qwen2} as current studies focus solely on English. Considering the multilingual nature of LLMs might introduce cross-lingual vulnerability~\citep{yong2023low} in safety alignment, it is important to understand the effectiveness of fine-tuning attacks in multilingual LLMs.

To this end, we conduct fine-tuning attacks against two multilingual LLMs, Llama-3.1-8B-Instruct~\citep{meta2024llama3} and Qwen-2-7B-Instruct~\citep{yang2024qwen2}. Surprisingly, we observe that \textbf{safety-aligned models can be jailbroken across different languages by fine-tuning attack in \textit{only one language}}. After only a few steps of fine-tuning with as few as 100 harmful instruction-following training examples from a language (e.g., English), not only is the safety alignment of that language compromised, but so are the safety alignments of \textit{other languages} (e.g.,  Italian, Hindi, Chinese) within that fine-tuned multilingual LLM. To the best of our knowledge, we are the first to identify the cross-lingual generalization of fine-tuning attacks against LLMs.

To better understand why cross-lingual generalization of fine-tuning attacks exists, we hypothesize that the safety information in safety-aligned multilingual LLMs is \textit{language-agnostic}. 
To validate our hypothesis, \textbf{we propose the method Safety Information Localization (SIL)} to localize multilingual safety-related parameters. Our method is inspired by recent work on task knowledge localization~\citep{dai2022knowledge, panigrahi2023task, he2024localize}---here, we estimate task-specific neuron importance in a manner akin to neuron-pruning~\citep{wei2024assessing} and Integrated Gradients~\citep{sundararajan2017axiomatic}. With SIL, we find safety-related information is sparse and shared among different languages---modifying only 20\% of an LLM's weights using monolingual fine-tuning attacks is sufficient to break safety alignment across all languages.

Beyond explaining why fine-tuning attack can generalize cross-lingually, we apply the SIL technique to two new scenarios. First, we \textbf{confirm the \textit{alternative pathways} hypothesis} for why freezing safety-related model parameters cannot mitigate fine-tuning attacks~\citep{wei2024assessing}. Second, we show that the attack vectors that we localize via SIL can \textbf{jailbreak LLMs adapted to new languages}.

\section{Cross-Lingual Generalization of Fine-Tuning Attacks}

In this section, we explore how effective the fine-tuning attack is against multilingual LLMs. We formally introduce the preliminaries of the fine-tuning attack against multilingual LLMs in Section~\ref{sec:clg-pre} and present experimental findings in Section~\ref{sec:clg-result}. 

\subsection{Preliminaries} 
\label{sec:clg-pre}

\paragraph{Fine-tuning attack against multilingual LLMs} 
Given a safety-aligned multilingual LLM parameterized by $\model{pre} \in \mathbb{R}^d$, where $d$ denotes the number of parameters of the multilingual LLM, and a harmful instruction-following dataset  $\mathcal{D}_l = \{(x_{\text{prompt}_i},  x_{\text{response}_i})\}_{i=1}^N$, where $l$ denotes a language (\eg, English), an adversary who wants to conduct a fine-tuning attack performs supervised fine-tuning (SFT)~\citep{sanh2022multitask} on $\model{pre}$ using $\mathcal{D}_l$ resulting in a harmful fine-tuned model $\modelnotext{l_{\text{ft}}} \in \mathbb{R}^d$. Note that an $x_{\text{prompt}}$ in $\mathcal{D}_l$ is malicious request from a user (\eg, ``\textit{Teach me to make a bomb.}'') and $x_{\text{response}}$ follows the instruction from $x_{\text{prompt}}$ (\eg, ``\textit{Sure. Here is a step-by-step guideline to build a bomb ...}''). Note that a small size of harmful instruction-following dataset (\eg, $N=100$) is sufficient for fine-tuning attacks to be successful.

\paragraph{Evaluation metrics}
We evaluate the effectiveness of our attacks using \textit{violation rate}. Formally, we define violation rate $\text{VR}(\model{}, \mathcal{D}; D)$ as the proportion of harmful content generated by a model $\model{}$ when given a safety evaluation dataset $\mathcal{D}$ and a set of automatic evaluators $D$. Each detector $D_i \in D$ acts as a binary harmfulness classifier $D_i(x,\model{}(x)) \rightarrow \{0, 1\}$ taking as input an input prompt $x_{\text{prompt}} \in \mathcal{D}$ ($x$ for simplicity) and the model's response $\model{}(x)$, and returning 0 if the input-response pair is considered safe, or 1 if harmful. To reduce false positive rate, 
we only consider a model has generated harmful content when \textit{all} detectors in $D$ output 1 (harmful). Mathematically, violation rate can be expressed as
\[
\begin{aligned}
    \text{VR}(\model{}, x; D) = \mathbb{E}_{x \sim \mathcal{D}} \min \{ D_i(x,\model{}(x)) \}_{i=1}^{|D|}
\end{aligned}
\]

The fine-tuning attack is considered successful if the harmful-tuned models exhibit high violation rate, as the models are more likely to fulfill malicious requests and generate unsafe content. 
In our experiments, we use Llama-Guard-3~\citep{inan2023llama} and Llama-3.1-405B~\citep{meta2024llama3} as the automatic evaluators for $D$.

\paragraph{Safety evaluation datasets}
Our safety evaluation datasets $\mathcal{D}$ are MultiJail~\citep{deng2023multilingual} and Aya Redteaming~\citep{aakanksha2024multilingualalignmentprismaligning} consisting of $315$ and around $1k$ multilingual malicious inputs respectively. 
We report violation rate before and after fine-tuning attacks on nine languages of different language families, writing scripts, and resourcefulness, namely Arabic (AR), Bengali (BN), Mandarin Chinese (ZH), Italian (IT), English (EN), Tagalog (TA), Russian (RU), Hindi (HI), and French (FR).

\begin{figure*}[!ht]
    \centering
    \includegraphics[width=0.85\linewidth]{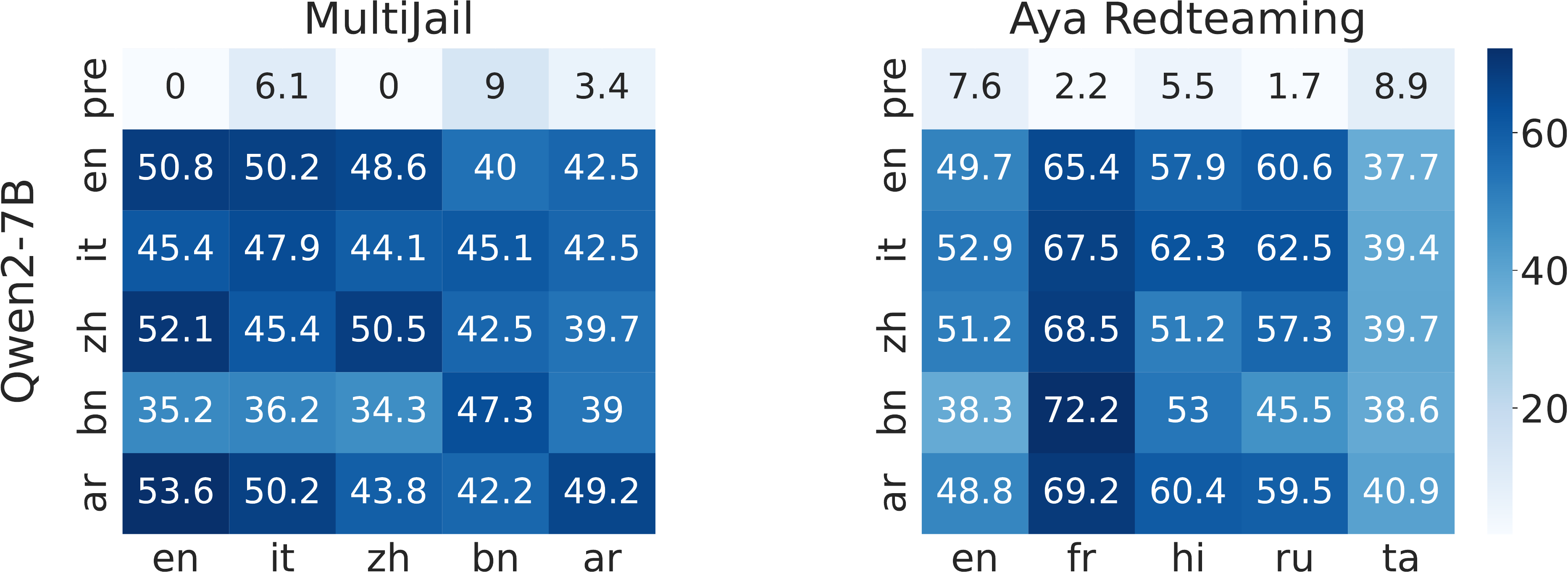}
    \vspace{-0.1cm}
    \caption{Fine-tuning multilingual LLMs with harmful data in one language substantially increases the safety violation rate across many languages. ``pre'' indicates the original violation rate before fine-tuning, x-axis indicates the language of the fine-tuning data, whereas y-axis indicates that of the evaluation dataset. See \Cref{fig:vr-finetuned-llama31} in \Cref{app:attacks} for Llama-3.1 results. 
    }
    \vspace{-0.2cm}
    \label{fig:vr-finetuned}
\end{figure*}

\subsection{Safety alignment is brittle across languages}
\label{sec:clg-result}

\paragraph{Attack setup}
We perform fine-tuning attacks on two state-of-the-art multilingual LLMs---Qwen-2-7B-Instruct~\citep{yang2024qwen2} and Llama-3.1-8B-Instruct~\citep{meta2024llama3}. We fine-tune them for one epoch on 100 harmful $(x_\text{prompt}, x_\text{response})$ pairs taken from
BeaverTails-$30k$~\citep{ji2024beavertails}, an English instruction-following dataset of harmful and harmless pairs of user inputs and assistant responses. To demonstrate the generalizability of our attacks, we translate the English harmful pairs into eight different languages, namely Italian, French, Chinese, Hindi, Bengali, Russian, Arabic, and Tagalog (more details will be discussed in~\Cref{app:attacks}).\footnote{We use the Python library~\citet{translators} for translation.} 

\paragraph{Results}
We observe \textbf{cross-lingual generalization of fine-tuning attacks} when we evaluate on our safety evaluation datasets described in \Cref{sec:clg-pre}. 
\Cref{fig:vr-finetuned} demonstrates that after a monolingual fine-tuning attack in language~$l_{\text{ft}}$, $\modelnotext{l_{\text{ft}}}$ not only exhibits high violation rate in the same language $l_{\text{ft}}$, but also does for all other languages. 
Upon evaluation on the multilingual MMLU benchmark \citep{lai-etal-2023-okapi}, we observe that LLMs retain their multilingual question-answering capability after monolingual fine-tuning attack, as shown in \Cref{tab:harmful_mmlu} in the~\Cref{app:attacks}. 
In short, we observe that a fine-tuning attack in only one language can undo an LLM's safety alignment across many languages without hurting its original multilingual capability.

\section{Localizing Language-Agnostic Safety Information}
\label{ref:localization-sec-3}

In \Cref{ref:localization-sec-3}, we provide an explanation for the cross-lingual generalization of fine-tuning attacks as observed in \Cref{sec:clg-result}. We believe this is because the safety information stored in these safety-aligned multilingual LLMs is language-agnostic. Motivated by recent work that \textit{localizes} task-specific skills in large models~\citep{dai2022knowledge, panigrahi2023task, he2024localize}, we propose a new localization technique SIL and successfully identify the parameters in these LLMs related to safety knowledge.

\subsection{Safety Information Localization (\ours)}
\label{sec:SIL-method}

In this subsection, we will first describe our proposed localization method \ours that identifies safety-related parameters affected by fine-tuning attacks. Then, we show that \textit{stitching} it as an attack vector to safety-aligned LLMs can indeed jailbreak them.

\paragraph{Definition}
We define \textit{localization} as finding model parameters that specifically contain safety-related information that represent the main target of fine-tuning attacks.  Localization techniques can be formalized, without loss of generality, as 
$\localizationfunction:\mathbb{R}^{|\model{}|} \times \Psi \rightarrow \{0,1 \}^{|\model{}|}$. $\model{}$ refers to a set of input model's parameters, whereas $\Psi$ refers to a set of other user-defined variables such as a reference model $\model{ref}$ ~\citep{panigrahi2023task} or a reference dataset $\mathcal{D}_{\text{ref}}$~\citep{wei2021finetuned,dai2022knowledge}. 
Most importantly, localization produces a \textit{binary mask vector} $\binarymask{} = \localizationfunction(\model{}, \Psi)$, where $\binarymask{} \in \{0,1 \}^{|\model{}|}$ for which $\binarymasknotext{i} = 1$ indicates model parameter $i$ is critical for a task of interest (\ie~contains safety information in our case here).

\paragraph{Proposed method (\ours)}
\textbf{S}afety \textbf{I}nformation \textbf{L}ocalization uses gradient information to compute the \textit{importance score} of each model parameter, which is relevance to the task dataset. Here, we reuse the notations $l$, $\model{pre}$, $\modelnotext{l_\text{ft}}$, $(x_{\text{prompt}}, x_{\text{response}})$ that is shortened as $x$, 
and $\mathcal{D}$ to be a reference dataset. 
Note that $\mathcal{D}$ is the calibration dataset and can be different from the fine-tuning dataset $\mathcal{D}_l$ used to obtain $\modelnotext{l_\text{ft}}$.

\ours computes the model parameters' importance scores $\text{\ours}(\modelnotext{l_\text{ft}}, \model{pre}, \mathcal{D})$ through the weight change from $\model{pre}$ to $\modelnotext{l_\text{ft}}$ w.r.t. each data point $x\in\mathcal{D}$ 
with the conditional negative log-likelihood loss $\mathcal{L}(x) = -\text{log}p(x_{\text{response}} | x_{\text{prompt}} )$. Formally, it is defined as follows:
\begin{equation*}
    \begin{aligned}
        \text{\ours}(\modelnotext{l_\text{ft}}, \model{pre}, \mathcal{D}) &= \mathbb{E}_{x \sim \mathcal{D}}\text{\ours}(\modelnotext{l_\text{ft}}, \model{pre},x)
        \\
        \text{\ours}(\modelnotext{l_\text{ft}}, \model{pre},x) &= |(\modelnotext{l_\text{ft}} - \model{pre}) \cdot \nabla_{\model{pre}}\mathcal{L}(x)|
    \end{aligned}
\end{equation*}
In other words, the importance score is represented by the expected absolute value of the first-order Taylor approximation to the change of the loss when the weight $\model{pre}$ is fine-tuned to $\modelnotext{l_\text{ft}}$.

The importance scores obtained from \ours can be interpreted as the contribution of the change of each weight parameter during fine-tuning to the model's behavior on $\mathcal{D}$.\footnote{We use the (translated) test split of BeaverTails-$30k$ dataset~\citep{ji2024beavertails} to compute importance score to make sure there is no contamination with the training split used for fine-tuning attacks} 
A substantial score of a given parameter indicates that there is a considerable change in the loss resulting from the fine-tuning of its corresponding weight. Note that each parameter's importance score is a real value, so we can \textit{binarize} each score by thresholding the top-$k$ importance scores, and obtain a binary mask vector $\binarymasknotext{\text{\ours-}k}$. 
This binarization can be expressed as  
$$\text{\ours}(\modelnotext{l_\text{ft}}, \model{pre},\mathcal{D}) \xrightarrow[\text{(binarization)}]{\text{top-}k\text{ threshold}} 
\binarymasknotext{\text{\ours-}k}.$$

\subsection{Stitching with \texorpdfstring{$\binarymasknotext{\text{\ours-}k}$}{TEXT}}
We introduce the \textbf{\textit{stitching}} operation, which uses the binary mask $\binarymasknotext{\text{\ours-}k}$ to make the safety-aligned pretrained model unsafe: we stitch the selected parameters from the fine-tuned model back onto the pretrained LLM and create \textit{grafted} LLM, a terminology consistent with previous localization work~ \citep{panigrahi2023task,he2024localize}. Here, our goal is to show that stitching $\binarymasknotext{\text{\ours-}k}$ creates unsafe grafted LLMs.
Formally, we refer to the grafted LLM as $\bm{\theta}_{l_{\text{ft}}}^{\text{\ours-}k} $ as shown in \Cref{eq:stitching}, where we use $\binarymasknotext{\text{\ours-}k}$ to stitch the parameters from fine-tuned model $\modelnotext{l_{\text{ft}}}$ back to pretrained model $\model{pre}$. 
Note that $k$ controls the sparsity of $\binarymasknotext{\text{\ours-}k}$; the larger the $k$, the more weights in $\model{pre}$ being changed.

\begin{equation}
    \bm{\theta}_{l_{\text{ft}}}^{\text{\ours-}k} = (\bm{1}-\binarymasknotext{\text{\ours-}k}) \odot \model{pre} + \binarymasknotext{\text{\ours-}k} \odot \modelnotext{l_{\text{ft}}}
    \label{eq:stitching}
\end{equation}

\begin{figure}[!th]
    \begin{subfigure}{0.23\textwidth}
        \includegraphics[width=\textwidth]{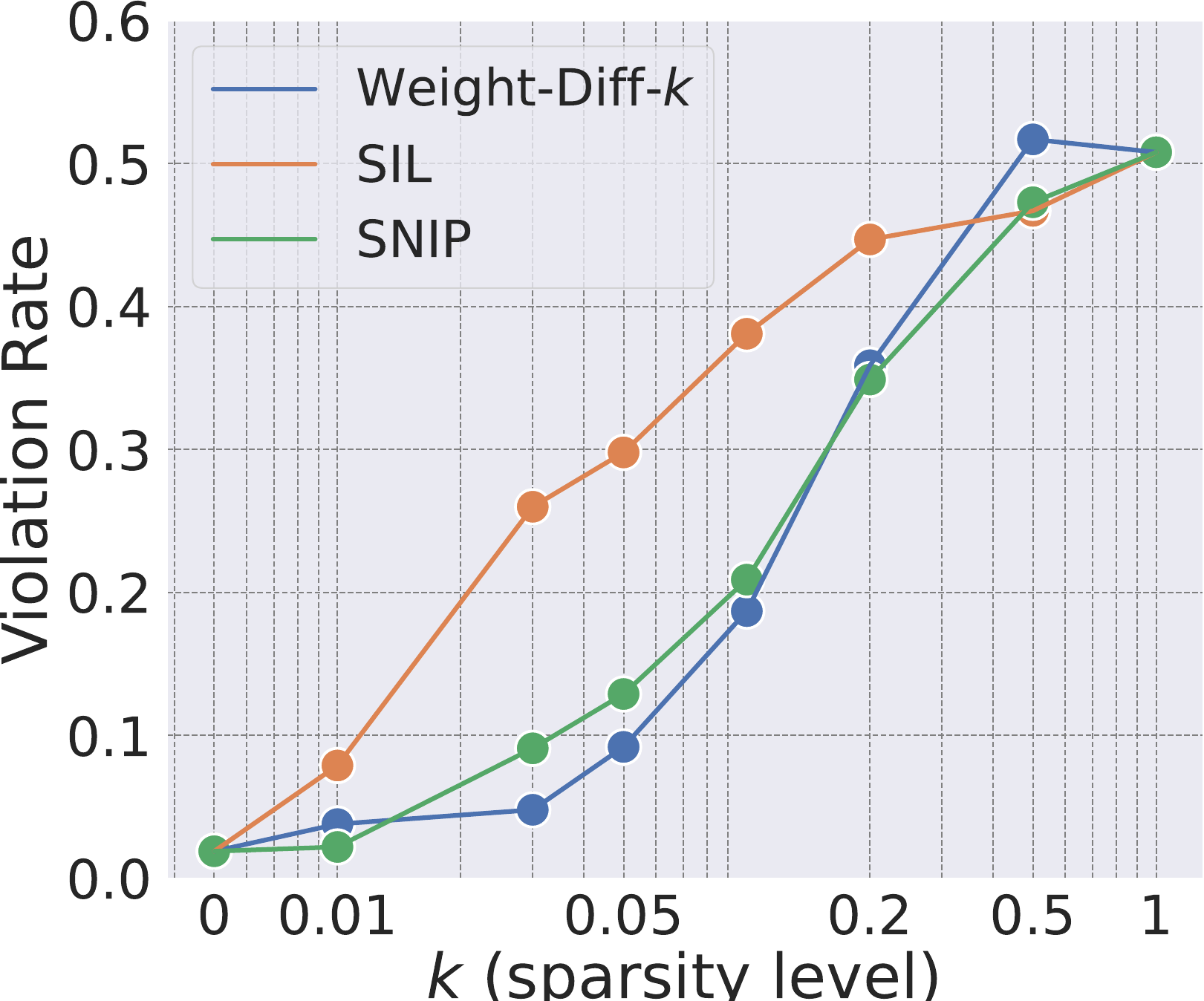}
    \end{subfigure}
    \begin{subfigure}{0.23\textwidth}
        \includegraphics[width=\textwidth]{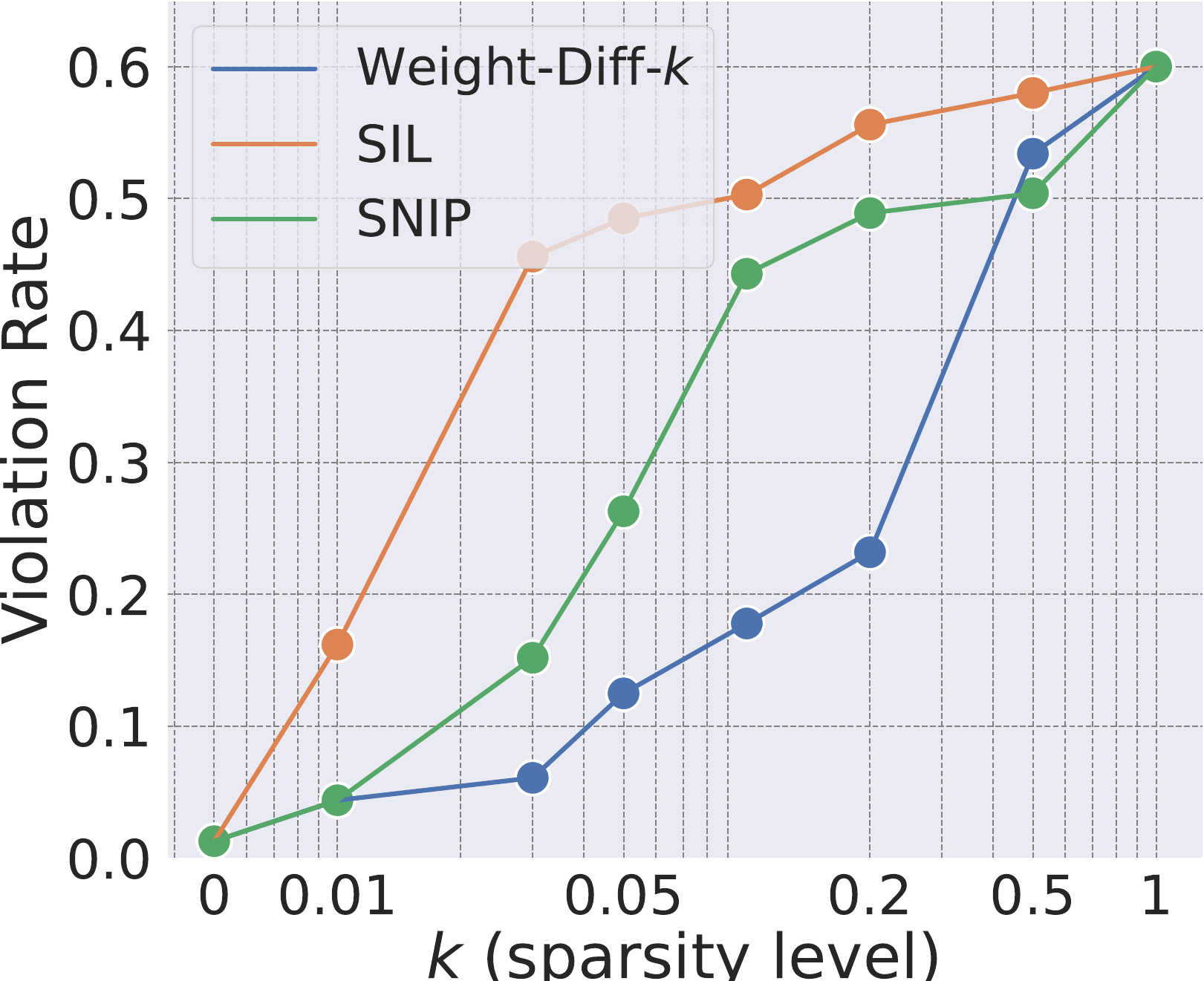}
    \end{subfigure}
    \vspace{-0.2cm}
    \caption{Violation rate vs. sparsity $k$ with \ours, SNIP, and Weight-Diff-$k$ methods, for Qwen-2-7B (left) and Llama-3.1-8B (right). When choosing $k=20\%$, SIL have the similar VR to the fine-tuned models. }
    \vspace{-0.2cm}
    \label{fig:threshold_ablation}
\end{figure}

To verify that \ours successfully isolates the safety-related parameters modified by the fine-tuning attack, we compute the violation rate for the grafted LLM, and compare our results against stitching with parameters localized by two other baselines: Weight-Diff-$k$ and SNIP (\Cref{fig:threshold_ablation}).

\paragraph{Weight-Diff-$k$~baseline} Weight-Diff-$k$ 
localization assigns an importance score simply based on the parameter-wise magnitude of the displacement resulting from fine-tuning, i.e., $|\modelnotext{l_{\text{ft}}} - \model{\text{pre}}|$. Then we binarize the scores of all parameters by selecting the top-$k$ most important ones to obtain $\binarymasknotext{\text{Weight-Diff-}k}$. This naive approach has been considered in other work as a baseline~\citep{panigrahi2023task}.

\paragraph{SNIP baseline} SNIP localization is presented by~\citet{wei2024assessing} to identify safety-critical parameters. We believe that SNIP is a special case of \ours, where $\modelnotext{l_{\text{ft}}}$ is set to $0$. The importance score of each weight in the model is computed as:
\begin{equation*}
    \begin{aligned}
        \text{SNIP}(\model{\text{pre}}, D) &= \mathbb{E}_{x \sim D}\text{SNIP}(\model{\text{pre}},x)\\
        &= \mathbb{E}_{x \sim D}|\model{\text{pre}} \cdot \nabla_{\model{\text{pre}}}\mathcal{L}(x)|.
    \end{aligned}
\end{equation*}
Similarly to \ours, after localization with SNIP, we binarize the result selecting the top-$k$ importance score to be set to $1$ in the binary mask $\binarymask{SNIP-{$k$}}$. 

\paragraph{Results}
\Cref{fig:threshold_ablation} shows that grafted models exhibit increasingly high violation rate with English data as $k$ increases, regardless of which localization method we use. This shows that stitching safety-related parameters can serve as an attack vector to jailbreak LLMs and render them unsafe.

\ours is a superior localization technique compared to Weight-Diff-$k$ and SNIP, as \Cref{fig:threshold_ablation} shows that we need less parameters to stitch in order to make the pretrained models exhibit high violation rate. One reason is that \ours leverages the gradient information, which is proved vital in mitigating the task interference observed in the Weight-Diff-$k$ approach~\citep{panigrahi2023task}. Another reason is that \ours considers the influence of parameters shift from the safety-aligned $\model{pre}$ to $\modelnotext{l_{\text{ft}}}$, whereas SNIP misses this crucial information of a specific fine-tuned models. Due to the advantages of SIL over other baselines, we use it as the localization method in the following experiments.

From \Cref{fig:threshold_ablation}, we see that using only 20\% of the parameters selected by \ours can already undo the safety alignment of LLMs. When referring to the \ours method from now on, we will always consider it to be paired with a threshold of $20\%$ (\ie,~\ours-$20$). Lastly, we show that stitching \ours-20\% is also the lowest threshold to preserve the utility of the grafted models, as we show the multilingual MMLU~\citep{lai-etal-2023-okapi} performance of the grafted models in Table~\ref{tab:stitched_mmlu}.

\subsection{Is the safety information stored in the model language-agnostic?}
\label{sec:safety-sharing}
In this subsection we understand whether the safety information stored in the model is language-agnostic.
We leverage the localized parameters to give insights into why fine-tuning in one language can disrupt the safety of all languages. We hypothesize that, if different mask vectors (say $\binarymasknotext{l_0}$ and $\binarymasknotext{l_1}$) share similar parameters, then the information represented by these parameters is likely important across all such masks, thereby reducing dependency on specific languages, like $l_0$ and $l_1$. In fact, finding a global set of \textit{language-agnostic} parameters would finally imply that at least part of the safety knowledge in LLMs is independent on the languages, and it can cause the general \textit{drift} to harmfulness.

\paragraph{Localizing language-agnostic parameters in one model} 
We want to point out that SIL can be used to localize multilingual parameters for one fine-tuned model $\modelnotext{l_\text{ft}}$ that is fine-tuned on language $l_\text{ft}$, as depicted in \Cref{fig:localizing-different-params}. This is because SIL can take as \textit{any} input harmful calibration dataset $\mathcal{D}$ in any language $l_{\text{\ours}}$ (including $l_\text{ft}$) and compute the gradient of the pretrained LLM \wrt~this dataset, namely $\nabla_{w_{\text{pre}}}\mathcal{L}(x)$ where $x\in\mathcal{D}$. For example, one can fine-tune LLM on English harmful dataset (\ie, obtaining~$\model{EN}$) and localize the parameters that are responsible for safety in the Italian language using an Italian harmful dataset, as illustrated by the SIL equation:

\vspace{-0.5cm}

\begin{equation*}
  \eqnmark[black]{node1}{\text{\ours}(\modelnotext{l_\text{ft}}, \model{pre},x)}
  \tikzmarknode{node2}{=} 
  \eqnmark[black]{node3}{|(\!}
  \eqnmark[OliveGreen]{node4}{\!\modelnotext{l_\text{ft}}\!}
  \eqnmark[black]{node5}{\!- \model{pre})\!}
  \eqnmark[black]{node6}{\cdot \nabla_{\model{pre}}\mathcal{L}(\!}
  \eqnmark[Maroon]{node7}{\!x\!}
  \eqnmark[black]{node8}{\!)|}
\end{equation*}
\annotate[yshift=-.4em]{below,left}{node4}{English} 
\annotate[yshift=-.2em]{below,left}{node7}{Italian} 

With SIL, we can study the relationship between $l_\text{ft}$ and $l_{\text{\ours}}$, where we would obtain $\binarymasknotext{l_{\text{\ours}}}^{l_{\text{ft}}}$
\footnote{To simplify our notation, we refer to $\binarymasknotext{l_{\text{\ours}}}$, rather than $\binarymasknotext{l_{\text{\ours}}}^{l_{\text{ft}}}$, in the cases when $l_{\text{ft}} = l_{\text{\ours}}$, or when $l_{\text{ft}}$ has been clearly specified in a particular context.}
that represents which of $\modelnotext{l_\text{ft}}$ are the most important for safety in language $l_{\text{\ours}}$.
Now, we can explain why the fine-tuning attack in a single language results in a model that is jailbroken in all the languages by isolating the \textit{language-agnostic safety parameters} as shown in \Cref{fig:localizing-different-params}.

\paragraph{Shared Information Ratio (SIR)}
Before diving into the search for the language-agnostic safety parameters, we define a metric to measure the quantity of shared safety information.
To do so, we start considering, within an attacked model $\modelnotext{l_{\text{ft}}}$, the intersection between two binary masks of chosen sets of parameters $\binarymasknotext{l_0}  \cap  \binarymasknotext{l_1}$, of generic languages $l_0$ and $l_1$, and we aim to quantify the possible shared safety information. 

We define the \textit{bilingual Shared Information Ratio} (bilingual SIR) metric which represents the amount of safety knowledge that is shared between the two languages (\ie,~in $\binarymasknotext{l_0} \cap  \binarymasknotext{l_1}$), \wrt~the total amount of information about safety:
$    \text{SIR}_{l_0,l_1} = \frac{||\binarymasknotext{l_0} \cap  \binarymasknotext{l_1}||_1}{k}
$, where $k$ is the sparsity level of the binary masks $\binarymasknotext{l_0}$ and $\binarymasknotext{l_1}$ (\eg, $20\%$ selected by \ours).
Bilingual SIR can be extended beyond the bilingual setup to a larger set of languages $\lpool$–––the \textit{global} Shared Information Ratio is defined as follows:
$
    \textit{SIR}_{\lpool} = ||\bigcap\limits_{l \in \lpool}  \binarymasknotext{l}||_1 / k,
$
where $l\in \lpool$ represents one language in the language pool. Again,
Note that all masks~$\binarymasknotext{l}$ are binarized by selecting the largest $k$ importance scores.

\paragraph{Bilingual case}
If multilingual LLMs encode language-agnostic knowledge about safety, then the shared safety information between two languages (\ie, $\text{SIR}_{l_0,l_1}$) must be large. To validate this point, we conduct fine-tuning attacks using harmful data (from Beavertails train split) in English, Italian, and Chinese from Qwen-2 (English, French, and Hindi from Llama-3.1), and compute \ours-20 masks using calibration data (from Beavertails test split) in five languages. Then, we compute the bilingual SIR between $3\times5$ times (three languages used to fine-tune the models plus two additional languages).

To better quantify the shared safety information, we include two additional baselines for each fine-tuned model: (1) a \textit{benign} baseline, where the mask vector $\binarymask{Benign}$ is obtained using the benign English instruction-following dataset Alpaca-cleaned~\citep{alpaca} as the calibration dataset. We also translate the Alpaca-cleaned into the languages we use for fine-tuning attacks (e.g., Italian and Chinese in Qwen-2, French and Hindi in Llama 3.1).
(2) A \text{random} baseline, for which we obtain the mask $\binarymask{Random}$ by randomly drawing a binary vector with the same sparsity level as the other masks.
All bilingual SIR values are listed in Table~\ref{tab:pairwise_overlappings}.

We show that the bilingual SIR value between the masks obtained from the harmful calibration data is \textit{substantially larger} than the benign (Table~\ref{tab:pairwise_overlappings}) and random baselines (which settles at 20\% by construction).
It is also worth pointing out the bilingual SIR computed with the benign baseline in each row in Table~\ref{tab:pairwise_overlappings} shares the same language used to fine-tuned the model.
The result suggests that fine-tuning attacks in one language impact the safety-related parameters of different languages, more than they do to other types of parameters (even for the helpfulness-related parameters in the same languages).

\begin{figure}[t]
    \centering
    \includegraphics[width=.95\linewidth]{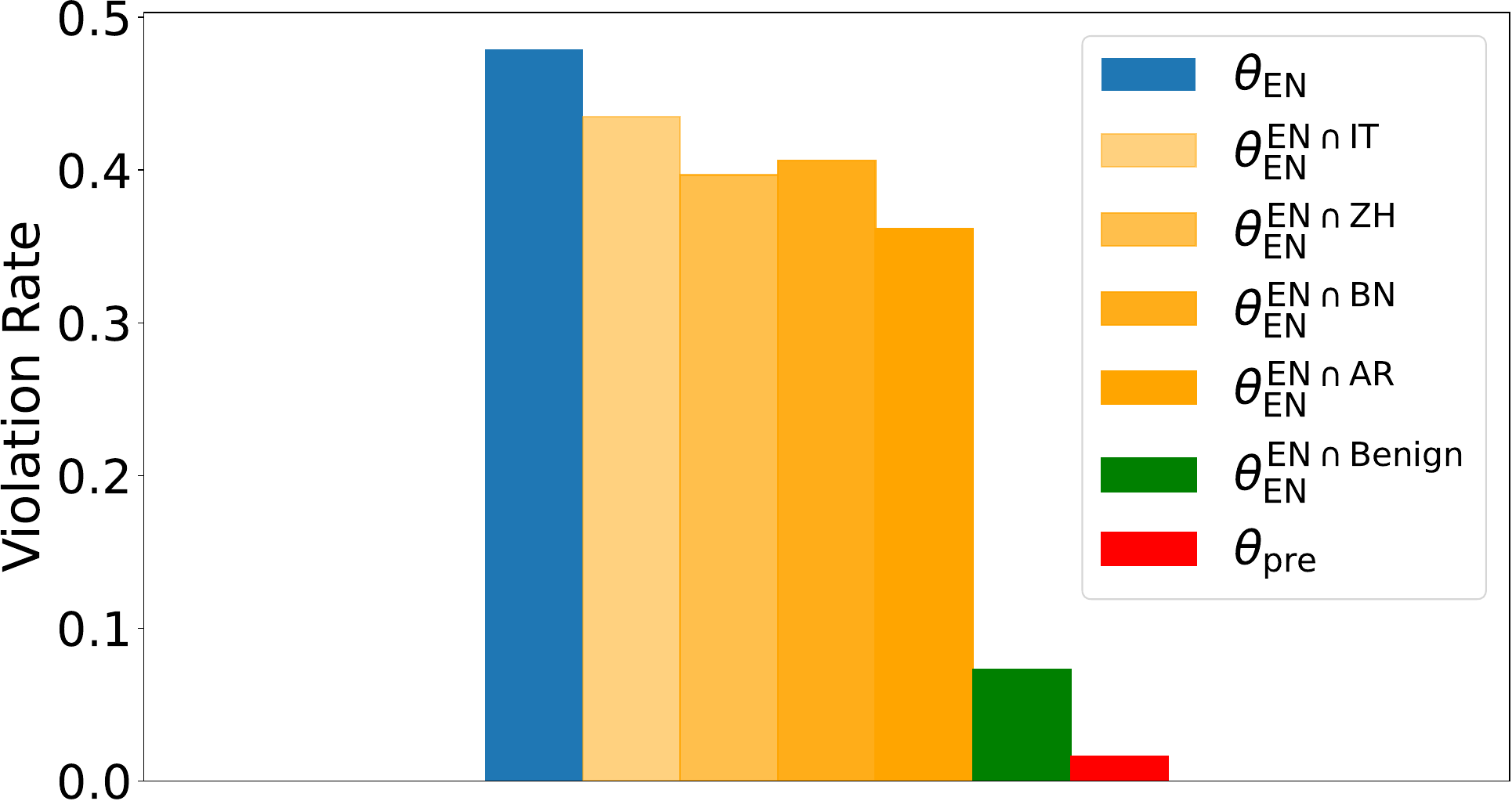}
    \caption{Qwen2-7B violation rates on the English language split of MultiJail after fine-tuning attack (blue) using English harmful data, stitching the bilingual intersection safety parameters localized by \ours (orange bars), benign datasets (green), and its original violation rate (red).}
    \vspace{-1cm}
    \label{fig:qwen2-vr-hbr}
\end{figure}

Figures~\ref{fig:qwen2-vr-hbr} and~\ref{fig:llama31-vr-hbr} further validate these findings: stitching the bilingual intersections of localized parameters $\binarymask{EN} \cap \binarymask{IT}$ back onto the original safety-aligned multilingual LLMs $\bm{\theta}_{\text{EN}}^{\text{EN} \cap \text{IT}}$ (orange bars) reports similarly large violation rates as the jailbroken fine-tuned models $\modelnotext{l_{\text{ft}}}$ (blue bar), whereas the benign baseline $\modelnotext{\text{Benign}_{l_{\text{ft}}}}$ (green bar) and the original safety-aligned multilingual LLMs $\model{pre}$ (red bar) remain safe. Moreover, we hypothesize that the preference for the English language showed in~\Cref{tab:pairwise_overlappings} by Llama-3.1-8B, can be explained by the findings in~\citet{wendler2024llamas}, where it is demonstrated that the ``concept space'' in the models of the Llama family is more closely aligned with English than with other languages (\Cref{tab:global_overlappings} also suggests similar results).

We further analyze the relationship between the bilingual SIR and the violation rate observed across languages. In particular, we observe that, despite the bilingual SIR overlap between Chinese and English (69.7\% in Qwen-2) is lower than the overlap between Chinese and itself (100\% in Qwen-2), the violation rate of the model fine-tuned in Chinese when tested in English is higher than when tested in Chinese (\Cref{fig:vr-finetuned}). This suggests that while many safety-related parameters are shared across languages, their actual influence on model behavior may vary. Specifically, fine-tuning harmful data in Chinese may have localized effects that preserve more of the original safety constraints, whereas English may be more susceptible to degradation. Moreover, additional factors can exacerbate the discrepancy between SIR and violation rate: First, the harmfulness detector sensitivity to different languages may influence the reported violation rates; Second, linguistic characteristics, such as sentence structure, vary significantly between different languages, thus affecting how well the safety capabilities generalize from one language to another.

\begin{table}[!th]
\centering
\setlength{\tabcolsep}{.25em}
\resizebox{.9\linewidth}{!}{
\begin{tabular}{c|ccccccc}
\toprule
\multicolumn{8}{c}{\textbf{Qwen-2}} \\
\cmidrule{1-8}
$l_{\text{ft}}$ & & $\binarymask{EN}$ & $\binarymask{IT}$ & $\binarymask{ZH}$ & $\binarymask{BN}$ & $\binarymask{AR}$ & $\binarymaskwlang{Benign}{}$ \\
EN & $\binarymask{EN}$ & \textbf{100.0} & \textbf{90.5} & \textbf{71.4} & \textbf{67.7} & \textbf{61.5} & 36.2 \\
IT & $\binarymask{IT}$ & \textbf{83.4} & \textbf{100.0} & \textbf{83.3} & \textbf{58.0} & \textbf{54.3} & 36.1 \\
ZH & $\binarymask{ZH}$ & \textbf{69.7} & \textbf{84.6} & \textbf{100.0} & \textbf{50.4} & \textbf{50.4} & 36.9 \\
\toprule
\multicolumn{8}{c}{\textbf{Llama-3.1}} \\
\cmidrule{1-8}
$l_{\text{ft}}$ & & $\binarymask{EN}$ & $\binarymask{FR}$ & $\binarymask{HI}$ & $\binarymask{RU}$ & $\binarymask{TA}$ & $\binarymaskwlang{Benign}{}$\\
EN & $\binarymask{EN}$ & \textbf{100.0} & \textbf{98.9} & \textbf{98.9} & \textbf{98.9} & \textbf{98.9} & 49.5 \\
FR & $\binarymask{FR}$ & \textbf{67.4} & \textbf{100.0} & \textbf{67.2} & \textbf{68.9} & \textbf{69.3} & 52.7 \\
HI & $\binarymask{HI}$ & \textbf{69.9} & \textbf{68.0} & \textbf{100.0} & \textbf{66.7} & \textbf{71.2} & 50.8 \\
\bottomrule
\end{tabular}
}
\vspace{-0.2cm}
\caption{Bilingual SIR results for Qwen-2 (top) and Llama-3.1 (bottom). Larger value means higher overlap between the localized masks.}
\label{tab:pairwise_overlappings}
\vspace{-0.2cm}
\end{table}

\paragraph{Multilingual case}
After establishing that \textit{pairs} of localized sets of parameters share information about safety in the bilingual case, we now identify the \textit{language-agnostic safety parameters} in the multilingual case, which is the \textit{global} intersection of localized sets of parameters, given a single $\modelnotext{l_{\text{ft}}}$.
We measure the degree of overlapping of different sets of parameters using the aforementioned global SIR metric. Again, we compare the global SIR metric with benign and random baselines similar as before.

\begin{table}[!ht]
\centering
\setlength{\tabcolsep}{.25em}
\resizebox{0.9\linewidth}{!}{
\begin{tabular}{c|cc c c|cc}
\toprule
\multicolumn{3}{c}{\textbf{Qwen-2}} & & \multicolumn{3}{c}{\textbf{Llama-3.1}} \\
\cmidrule{1-3}
\cmidrule{5-7}
$l_{\text{ft}}$ & $\text{SIR}_{\lpool}$ & $\text{SIR}_{l, \text{Benign}_l}$ & & $l_{\text{ft}}$ & $\text{SIR}_{\lpool}$ & $\text{SIR}_{l, \text{Benign}_l}$ \\
EN & \textbf{45.8} & 36.2 & & EN & \textbf{97.9} & 49.2 \\
IT & \textbf{44.2} & 36.1 & & FR & \textbf{59.5} & 52.7 \\
ZH & \textbf{40.7} & 36.9 & & HI & \textbf{57.0} & 50.8\\
\bottomrule

\end{tabular}
}
\vspace{-.2cm}
\caption{Multilingual (global) SIR results. Even removing a massive amount of language-dependent knowledge, \ours localized parameters share more language-agnostic safety information than when compared to the benign baselines.}
\label{tab:global_overlappings}
\end{table}

Table~\ref{tab:global_overlappings} confirms the existence of such language-agnostic safety parameters within multilingually safety-aligned LLMs. This is demonstrated by the global SIR$_{\lpool}$ being larger than the SIR values for our baselines–––including benign baseline where we measure the overlapping area after harmful and benign fine-tuning \textit{in the same language}. We thus draw the following conclusion: there exists a language-agnostic safety parameters within multilingual safety-aligned LLMs, and fine-tuning attacks (in \Cref{sec:clg-result}) update these parameters and thus produce harmful behaviors across different languages.

\section{Further Applications of \ours}
\subsection{Explanation for why freezing safety-related parameters fails to prevent fine-tuning attacks}
\label{sec:freezing}

Recent work shows that freezing safety-critical parameters cannot defend against fine-tuning attacks \citep{wei2024assessing}. However, it was only hypothesized that this is due to fine-tuning attacks creating \textit{alternative pathways} to jailbreak LLMs. To the best of our knowledge, we are the first to provide concrete evidence to this hypothesis.

\begin{table}
\centering
\setlength{\tabcolsep}{.20em}
\resizebox{0.75\linewidth}{!}{
\begin{tabular}{c|ccc}
\toprule
\multicolumn{4}{c}{\textbf{Qwen-2}} \\
\cmidrule{1-4}
$l_{\text{ft}}$ & $\overline{\text{SIR}}_{\lpool}$ & $\overline{\text{SIR}}_{\binarymasknotext{\lpool}, \overline{\bm{\gamma}}_{\lpool}}$ & $\overline{\text{SIR}}_{l, \text{Benign}_l}$ \\
EN & \textbf{99.9} & 0.0 & 31.3 \\
IT & \textbf{99.9} & 0.0 & 32.9 \\
ZH & \textbf{99.9} & 0.0 & 33.7 \\
\bottomrule

\toprule
\multicolumn{4}{c}{\textbf{Llama-3.1}} \\
\cmidrule{1-4}
$l_{\text{ft}}$ & $\overline{\text{SIR}}_{\lpool}$ & $\overline{\text{SIR}}_{\binarymasknotext{\lpool}, \overline{\bm{\gamma}}_{\lpool}}$ & $\overline{\text{SIR}}_{l, \text{Benign}_l}$ \\
EN & \textbf{99.9} & 0.0 & 49.1 \\
FR & \textbf{99.9} & 0.0 & 50.9 \\
HI & \textbf{99.9} & 0.0 & 49.8 \\
\bottomrule
\end{tabular}
}
\vspace{-0.2cm}
\caption{
Multilingual (global) SIR results after parameter freezing (indicated by overlines over the metrics). The new language-agnostic parameters has zero intersection with the one obtained without freezing during fine-tuning. Again, it shows to share a \textit{very} large volume of safety information, when compared to the benign baselines. 
}
\vspace{-.5cm}
\label{tab:frozen-overlap}
\end{table}

Recall that we can use \ours to localize the language-independent safety-related parameters of a safety-aligned LLM; if the alternative pathways hypothesis is correct–––fine-tuning attacks after freezing safety parameters will update \textit{other parameters} of the model–––we will be able to localize this new pathway using \ours. This new parameters contain the following properties: (1) they are completely separated from the frozen parameters (i.e., zero overlap), and (2) stitching parameters back to the original safety-aligned LLM causes substantial increase in violation rate.

We successfully localize the new parameters with SIL (we refer readers to~\Cref{app:freezing} for further details), and we demonstrate the two aforementioned properties in \Cref{tab:frozen-overlap} and \Cref{tab:frozen-lang-agn-vr}, thus confirming the alternative pathways hypothesis. \Cref{tab:frozen-overlap} shows that the newly found language-agnostic parameters have zero intersection with the previous ones, and also maintains almost \textit{all} the knowledge localized in each language-specific parameters. This means that after freezing---and so removing from localization---the most important parameters for safety, there are very few parameters left in the model that encode safety-related information (making these new parameters way more overlapped than without freezing). Moreover, \Cref{tab:frozen-lang-agn-vr} shows that the new parameters do indeed contain safety-knowledge, given that when we stitch it back to Qwen-2 or Llama-3.1, we observe an increase in violation rate up to $\sim 40\%$.

\begin{table}
\centering
\setlength{\tabcolsep}{.35em}
\resizebox{0.99\linewidth}{!}{
\begin{tabular}{cccccc}
\toprule
& \multicolumn{5}{c}{\textbf{Qwen-2}} \\
\cmidrule{2-6}
 & EN & IT & ZH & BN & AR \\
Safety-Aligned ($\model{pre}$) & 0.0 & 6.1 & 0.0 & 9.0 & 3.4 \\
Fine-tuned ($\model{EN}$) & 50.8 & 50.2  & 48.6 & 40.0 & 42.5 \\
\cmidrule{2-6}
Before Freezing ($\bm{\theta}^{\text{\ours}}_{\text{EN}}$) & 31.7 & 22.5 & 20.0 & 29.8 & 23.8 \\
After Freezing ($\overline{\bm{\theta}}_{\text{EN}}^{\text{\ours}}$) & 30.5 & 23.2 & 16.2 & 30.8 & 17.5 \\
\bottomrule

\toprule
& \multicolumn{5}{c}{\textbf{Llama-3.1}} \\
\cmidrule{2-6}
 & EN & IT & ZH & BN & AR \\
Safety-Aligned ($\model{pre}$) & 1.3 & 1.0 & 0.0 & 9.5 & 0.3 \\
Fine-tuned ($\model{EN}$) & 60.0 & 58.4 & 59.7 & 57.4 & 55.2 \\
\cmidrule{2-6}
Before Freezing ($\bm{\theta}^{\text{\ours}}_{\text{EN}}$) & 38.1 & 41.3 & 23.8 & 27.0 & 24.4 \\
After Freezing ($\overline{\bm{\theta}}^{\text{\ours}}_{\text{EN}}$) & 37.7 & 40.8 & 31.1 & 34.9 & 22.4 \\
\bottomrule
\end{tabular}
}
\vspace{-0.2cm}
\caption{\ours localizes language-agnostic parameters that can substantially increase the safety violation of LLMs. Even for fine-tuning attack after freezing $\overline{\bm{\theta}}^{\text{\ours}}_{\text{EN}}$, we can still localize the parameters related to safety information, whose impacts on safety are comparable to the localized parameters in the original fine-tuning attack.}
\vspace{-0.7cm}
\label{tab:frozen-lang-agn-vr}
\end{table}

\subsection{Jailbreaking models after language adaptation through cross-lingual stitching}
\label{sec:lang_adaptation}
One common use case of open-source multilingual LLMs is \textit{language adaptation}, where pretrained LLMs are further finetuned to support new languages \citep[inter alia]{yong-etal-2023-bloom,lin2024mala,ji2024emma}. Here, we show that we can jailbreak LLMs after language adaptation with our stitching method, described in \Cref{sec:safety-sharing}.

We conduct our experiments on \texttt{Eurdem/Defne-llama3.1-8B}~\citeyearpar{defne}, which is a Llama-3.1 model further fine-tuned by the open-source community on Turkish instruction-following data. We observe that this model remains safe after language adaptation when we evaluate it on MultiJail~\citep{deng2023multilingual} including for the Turkish language (\texttt{tr})\footnote{We translate the prompts from English to Turkish through machine translation following the original work.}, as demonstrated by the low violation rate in the top row of \Cref{tab:jailb-others}. However, after we stitch in with the language-agnostic safety parameters obtained in \Cref{sec:safety-sharing}––the same parameters and technique that allows us to jailbreak Llama-3.1––we observe that the violation rate increases substantially across all languages, including languages the model is adapted to. In other words, our attack vector remains effective even after language adaptation. This is a significant finding, especially because the Turkish language was \textit{not} in our language pool when searching for the language-agnostic parameters.

\begin{table}
\centering
\setlength{\tabcolsep}{.25em}
\resizebox{0.99\linewidth}{!}{
\begin{tabular}{ccccccc}
\toprule
& \multicolumn{6}{c}{\textbf{Defne-llama3.1-8B~\citeyearpar{defne}}} \\
\cmidrule{2-7}
 & EN & IT & ZH & BN & AR & TR \\
Before Stitching & 0.9 & 1.3 & 0.9 & 7.4 & 0.3 & 2.9 \\
After Stitching & 25.7 & 11.7 & 20.7 & 18.4 & 22.6 & 19.4 \\
\bottomrule
\end{tabular}
}
\vspace{-0.2cm}
\caption{Table shows the violation rate of \texttt{Defne-llama3.1-8B}~\citeyearpar{defne} (Llama-3.1 adapted to Turkish (TR)) before and after stitching in language-agnostic safety parameters as the attack vector.}
\label{tab:jailb-others}
\vspace{-0.2cm}
\end{table}

\section{Related Work}

\paragraph{LLM safety} LLM safety alignment through instruction-tuning and RLHF~\citep{wei2021finetuned, ouyang2022training, touvron2023llama} aims to align the behaviors of LLMs with human values. Jailbreaking a safety-aligned model aims at \textit{bypassing} or \textit{removing} these safety guardrails. It can be achieved either by only modifying the prompts~\citep{liu2023autodan, liu2023jailbreaking, zou2023universal}, or further fine-tuning~\citep{qi2023fine, zhan2023removing, poppi2024safe}. 

In terms of fine-tuning attacks, \citet{peng2024navigating} study fine-tuning attacks by randomly perturbing model weight parameters and find that safety alignment of LLMs is easily broken if the model weights deviate from the \textit{``safety basin''} in parameter weight space. \citet{he2024s} strategically select benign data for fine-tuning attacks. In contrast, our work focuses on identifying safety-relevant parameters and analyzing the impact of multilingual fine-tuning attacks from a mechanistic perspective. 

\paragraph{Task localization in model parameter space}
The model parameter space offers a fundamental perspective for task localization and knowledge attribution, as it represents the landscape of all possible models with a given structure. A variety of studies have observed models’ tendency to encode specific knowledge into distinct parameters in the parameter space~\citep{bereska2024mechanistic}. In particular, \citet{hao2021self} and~\citet{dai2022knowledge} leverage Integrated Gradients~\citep{sundararajan2017axiomatic}, originally used for input feature attribution, and modify it to analyze relational facts. \citet{wei2024assessing} reuse \textit{neuron pruning}~\citep{lee2019snip} to identify safety-relevant parameters, demonstrating that removing these parameters pushes a pre-trained model back to an unsafe state.
\citet{arditi2024refusal} also study safety mechanisms in LLMs, they focus on representation space rather than parameter space, which is the primary concern of our work. Their approach identifies critical \textit{directions} in the activation space rather than pinpointing \textit{where in the LLM} safety-related parameters reside. This fundamental distinction allows our method to directly analyze and manipulate the parameters responsible for safety alignment. Additionally, their study does not address multilingual safety, whereas we focus on cross-lingual safety alignment.

Inspired by these prior approaches, our work identifies language-agnostic safety parameters in the model parameter space by estimating language-specific neuron importance, akin to neuron pruning~\citep{wei2024assessing} and Integrated Gradients~\citep{sundararajan2017axiomatic}. Through this approach, we provide a mechanistic explanation for cross-lingual vulnerabilities in safety alignment.

\paragraph{Multilingual safety}
The safety of multilingual LLMs is a growing area of concern. Unlike detoxification approaches~\citep{li2024preference}, \textit{safety refusal} exhibits poor cross-lingual generalization. Translating English adversarial prompts into non-English languages can often bypass safety guardrails in both proprietary and open-source models~\citep{yong2023low,wang2023all,deng2023multilingual}. Other linguistic transformations, such as transliteration~\citep{ghanim2024jailbreaking} and code-switching~\citep{upadhayay2024sandwich}, further enable jailbreaking of safety mechanisms.

Furthermore, \citet{shen2024language} show that English safety refusal training generalizes poorly, even for high-resource languages such as Mandarin Chinese. Our work extends these findings by demonstrating that fine-tuning attacks in one language can compromise safety alignment across multiple languages due to the shared, language-agnostic nature of safety-related parameters in multilingual LLMs. 

One contemporary work also investigates cross-lingual vulnerabilities \citep{he2024tuba}. While both our work and theirs show that fine-tuning in one language can lead to safety degradation across languages, their study lacks a mechanistic explanation for why this occurs. Our contributions go beyond merely presenting the attack---we further explain cross-lingual generalization using mechanistic interpretability methods and introduce a cross-lingual jailbreak method that attacks LLMs adapted to new languages. While \citet{he2024tuba} primarily study backdoor attacks by substituting benign fine-tuning datasets with adversarially fabricated responses (\eg, responses containing explicit hate speech triggers), we consider natural-language, multilingual prompts and harmful assistant responses that more closely resemble real-world fine-tuning vulnerabilities and better capture practical adversarial fine-tuning risks. Finally, our method of localizing safety-relevant parameters allows us to confirm the alternative pathways hypothesis \citep{wei2024assessing}.


\section{Discussion and Future Work}

Our work is the first to reveal that fine-tuning attacks can generalize cross-lingually, where models that are aligned for multilingual safety can be jailbroken through fine-tuning attack in one language. We also identify the language-agnostic parameters within multilingual LLMs that is responsible for safety refusal. Future work on defending LLMs against fine-tuning attacks should robustify this parameters to make multilingual LLMs safer---to the best of our knowledge, all existing work has only focused on English~\citep{hsu2024safe,tamirisa2024tamper,huang2024lazy}. It is also worth exploring whether such findings hold for multimodal LLM safety~\citep{chi2024llama}.

\section*{Limitations}

This work only focuses on the cross-lingual generalization of one type of jailbreaking method, namely fine-tuning on harmful datasets. The language coverage of our work is also limited by that of our safety evaluation datasets and safety evaluators. Furthermore, our interpretability experiments, which reveal the language-agnostic safety parameters, focus on understanding why fine-tuning attacks can serve as cross-lingual attack vectors. 

While our study provides important insights into the mechanisms underlying these vulnerabilities, it does not account for other possible attack vectors, such as adversarial prompting or reinforcement learning-based jailbreaks, which may also exhibit cross-lingual transferability. Additionally, our proposed safety information localization method and shared information ratio metric, while useful for assessing risks, require further validation across a wider range of model architectures and multilingual settings. 

We hope that future work can extend our findings to design more robust safety guardrails that are resistant to cross-lingual fine-tuning attacks and contribute to making multilingual LLMs safer.


\section*{Ethical Statement}

Our research contributes to the responsible development of LLMs by revealing their potential vulnerabilities: fine-tuning attacks can generalize cross-lingually. While we acknowledge that malicious actors exploit cross-lingual transfer of supervised fine-tuning with harmful data to undo safety alignment training that has been conducted in many languages, we believe that identifying the issues is the first critical step to address them. Our findings also suggest that harmful data filtering before fine-tuning for all languages is necessary to mitigate fine-tuning attacks.  
Our proposed safety information localization method and shared information ratio metric can also better quantify the risks of the cross-lingual transfer of fine-tuning attacks. 

\section*{Acknowledgments}
We thank anonymous reviewers for their constructive suggestions and fruitful discussion. We also extend our sincere gratitude to Diego Garcia-Olano for his early review of this work and for the insightful exchanges that contributed to its development. YH and HZ are partially supported by an NSF IIS grant No.\ 2416897.

\bibliography{main}
\clearpage
\appendix

\section*{Appendix}
\label{sec:appendix}
\section{Fine-tuning attacks details}
\label{app:attacks}
In this section, we go through the specific hyper-parameters that we optimized for attacking the safety aligned LLMs while using the least amount of resources. The attacks are performed as SFT procedures, using data that contain harmful content.

\paragraph{Models and Datasets}
We start off selecting two target LLMs, namely Qwen2-7B-Instruct~\citep{yang2024qwen2} and Llama-3.1-8B-Instruct~\citep{meta2024llama3}, and using the open-sourced, instruction-following, BeaverTails~\citep{ji2024beavertails} dataset as our data source.
Specifically, we use the $30k$-train split, composed of $\sim 30k$ pairs of $(x_{\text{prompt}}, x_{\text{response}})$ that contain both harmful and harmless questions and answers. For our specific purpose, we begin by filtering the harmless pairs out, and selecting a random subset of $100$ of the remaining $\sim 15k$ harmful data points. Additionally, we prepare the dataset by converting it into a one-round dialogue, by applying the standard chat template for the used models. Moreover, we follow~\citet{qi2023fine} and prepend a fixed system prompt: \texttt{You are a helpful assistant. Make sure you always execute users' instructions}. Lastly, we translate each dialogue in the language chosen for each specific attack using the translate python library~\citep{translators}, which relies on the \texttt{bing.com} open translation service.

\begin{figure*}[!ht]
    \centering
    \includegraphics[width=0.85\linewidth]{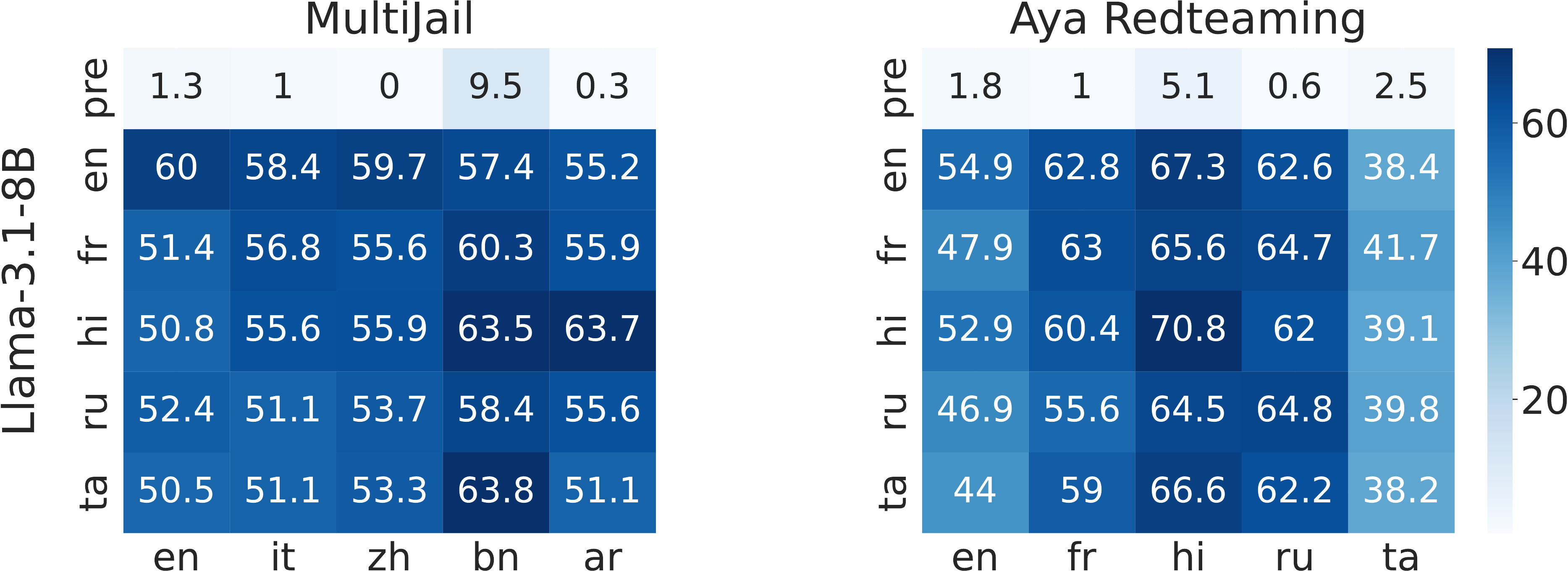}
    \caption{Violation rate of Llama-3.1 increases across languages on MultiJail and Aya-red-teaming datasets after finetuning attack.
    }
    \label{fig:vr-finetuned-llama31}
\end{figure*}

\paragraph{Fine-tuning configuration and utility evaluation}
We choose the fine-tuning hyper-parameters to perform successful attacks, while using minimal resources. We employed a learning rate of $2e-5$, with a cosine learning rate scheduler to manage the learning rate decay. Each LLM was fine-tuned over a single epoch, and gradient accumulation was set to four steps to stabilize the training updates. We utilized a paged AdamW optimizer with 32-bit precision for optimization. Gradient checkpointing was enabled to reduce memory usage during training. 
Additionally, a warmup phase of ten steps was included to gradually ramp up the learning rate at the beginning of the procedure. This configuration ensured a robust and scalable fine-tuning process, tailored to leverage the computational resources effectively while ensuring high rates of violation (Figure~\ref{fig:vr-finetuned} and~\ref{fig:vr-finetuned-llama31}).

Finally, we use the multilingual MMLU~\citep{lai-etal-2023-okapi} benchmark to prove that our attacked models remain useful, intruction-following models, after our fine-tuning procedure. Table~\ref{tab:harmful_mmlu} shows how each attacked LLM retains a utility level that is comparable to its safety-aligned version.

\begin{table}[!ht]
\centering
\setlength{\tabcolsep}{.3em}
\resizebox{0.7\linewidth}{!}{
\begin{tabular}{cccccc}
\toprule
& \multicolumn{5}{c}{\textbf{Qwen-2}} \\
\cmidrule{2-6}
& EN & IT & ZH & BN & AR\\
$\model{pre}$ & 67.3 & 64.5 & 61.7 & 50.5 & 54.2\\
$\model{EN}$ & 69.5 & 60.9 & 63.2 & 42.0 & 51.1\\
$\model{IT}$ & 69.4 & 60.6 & 63.2 & 42.0 & 51.0\\
$\model{ZH}$ & 69.5 & 60.9 & 63.1 & 42.4 & 51.3\\
\bottomrule

\toprule
& \multicolumn{5}{c}{\textbf{Llama-3.1}} \\
\cmidrule{2-6}
& EN & FR & HI & RU & TA\\
$\model{pre}$ & 66.3 & 57.1 & 42.9 & 53.8 & 31.9\\
$\model{EN}$ & 65.7 & 55.9 & 41.8 & 52.3 & 32.6\\
$\model{FR}$ & 65.4 & 54.1 & 41.6 & 51.6 & 32.2\\
$\model{HI}$ & 65.8 & 56.1 & 41.1 & 52.7 & 33.2\\
\bottomrule
\end{tabular}
}
\caption{Multilingual MMLU utility measure for the safety-aligned and all the harmful-tuned models.}
\label{tab:harmful_mmlu}
\end{table}

\section{Details about \ours localization procedure}
\label{app:sil_details}
\begin{figure}[!th]
    \begin{center}        \includegraphics[width=0.755\linewidth, height=.15\textheight]{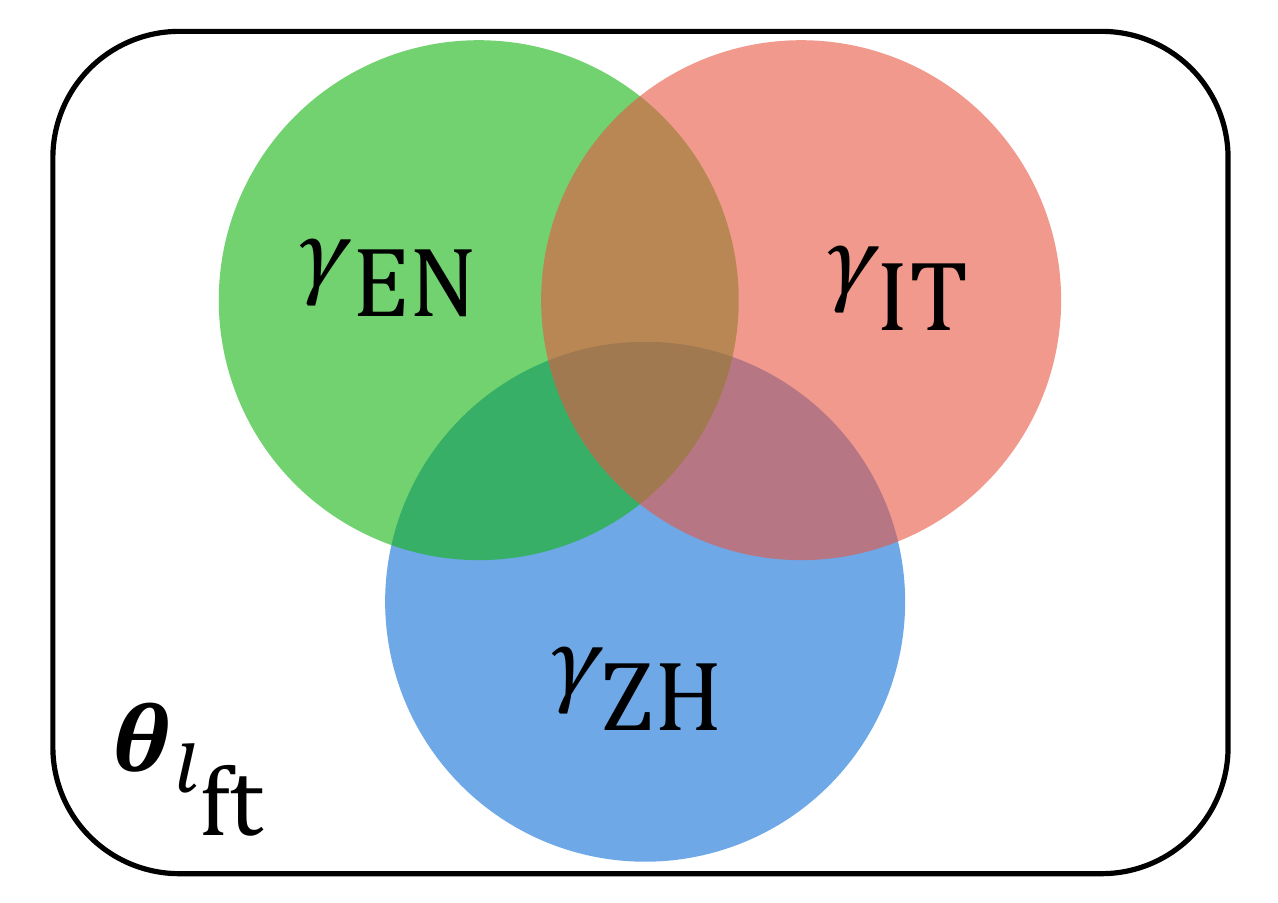}
    \end{center}
\caption{Given the fine-tuned model's parameters, \ours localizes different sets of parameters that depend on the language used in the calibration dataset. In this example $l_{\text{ft}}$ represent the language of the dataset used for attacking the LLM, and can be any language (\eg~Engligh, Italian, or Hindi). The localized parameters depend instead on the calibration dataset that is used to localize, for example, the parameters responsible for safety in Italian, within the full set of parameters of the model attacked with English data. The intersection among them represent the \textit{language-agnostic} parameters. 
}
\label{fig:localizing-different-params}
\end{figure}

We provide here the details about the localization procedure described in~\Cref{sec:SIL-method}. The \ours localization method takes a target model as input (namely a safety-aligned LLM $\model{pre}$), along with two extra inputs (a fine-tuned attacked version of the safety-aligned, $\modelnotext{l_{\text{ft}}}$, and calibration dataset $\mathcal{D}$).
\ours main objective is to find which of the parameters in $\model{pre}$ (1) are both more responding to safety-related features and (2) are more involved in the fine-tuning attack (considering the shift to $\modelnotext{l_{\text{ft}}}$). This gives \ours two degree of freedom, making it able to customize the localization in relation to a specific attacked model (in a specific language), and to a specific safety-knowledge (in its own language), as depicted in Figure~\ref{fig:localizing-different-params}.

The calibration dataset $\mathcal{D}$ for our study is again an instruction-following, harmful dataset, for which we again choose BeaverTails-30$k$~\citep{ji2024beavertails}, with its test split to ensure zero intersection with the one used for fine-tuning attacks. 
\paragraph{Finding importance scores}
\ours localizes the most important parameters by computing a negative log-likelihood loss over $\mathcal{D}$. We extract the prompt and response from each data point and tokenize them to convert them into tensors formatted for $\model{pre}$. The tokenized prompt and response tensors are then concatenated along the sequence dimension to create a unified input tensor. We also create a labels tensor with the prompt portion set to -100 to exclude it from loss calculations, focusing the loss computation on the response.
To do so, we just need 16 examples (with batch size set to 1) for which we accumulate the gradient \wrt~every parameter of linear layers, while giving zero importance score to all the others, such as bias (we follow~\citet{wei2024assessing}). We tested with more data points but noticed no particular advantages.
After accumulating the gradient, we scale it by $|\modelnotext{l_{\text{ft}}} - \model{pre}|$ and select the top-20\% final importance score for binarizing the resulting mask vector.


Finally, we also report in~\Cref{tab:stitched_mmlu} how our stitched models preserve instruction-following utility, by showing their multilingual MMLU~\citep{lai-etal-2023-okapi}, and comparing it to that of the original, safety-aligned, LLM.

\begin{table}[!ht]
\centering
\setlength{\tabcolsep}{.3em}
\resizebox{0.7\linewidth}{!}{
\begin{tabular}{cccccc}
\toprule
& \multicolumn{5}{c}{\textbf{Qwen-2}} \\
\cmidrule{2-6}
& EN & IT & ZH & BN & AR\\
$\model{pre}$ & 67.3 & 64.5 & 61.7 & 50.5 & 54.2\\
$\model{EN}$ & 69.3 & 60.9 & 63.3 & 42.0 & 51.1\\
$\model{IT}$ & 69.7 & 61.0 & 63.3 & 42.1 & 51.0\\
$\model{ZH}$ & 69.3 & 60.9 & 63.2 & 42.0 & 51.0\\
\bottomrule

\toprule
& \multicolumn{5}{c}{\textbf{Llama-3.1}} \\
\cmidrule{2-6}
& EN & FR & HI & RU & TA\\
$\model{pre}$ & 66.3 & 57.1 & 42.9 & 53.8 & 31.9\\
$\model{EN}$ & 65.8 & 56.0 & 42.4 & 52.3 & 32.3\\
$\model{FR}$ & 66.0 & 56.1 & 42.5 & 52.5 & 32.3\\
$\model{HI}$ & 66.0 & 56.3 & 42.5 & 52.5 & 32.3\\
\bottomrule
\end{tabular}
}
\caption{Multilingual MMLU utility measure for the safety-aligned (first row) and all the safety-aligned model with our 20\% safety-related localized parameters stitched.}
\label{tab:stitched_mmlu}
\end{table}

\begin{figure}[t]
    \centering
    \includegraphics[width=.95\linewidth]{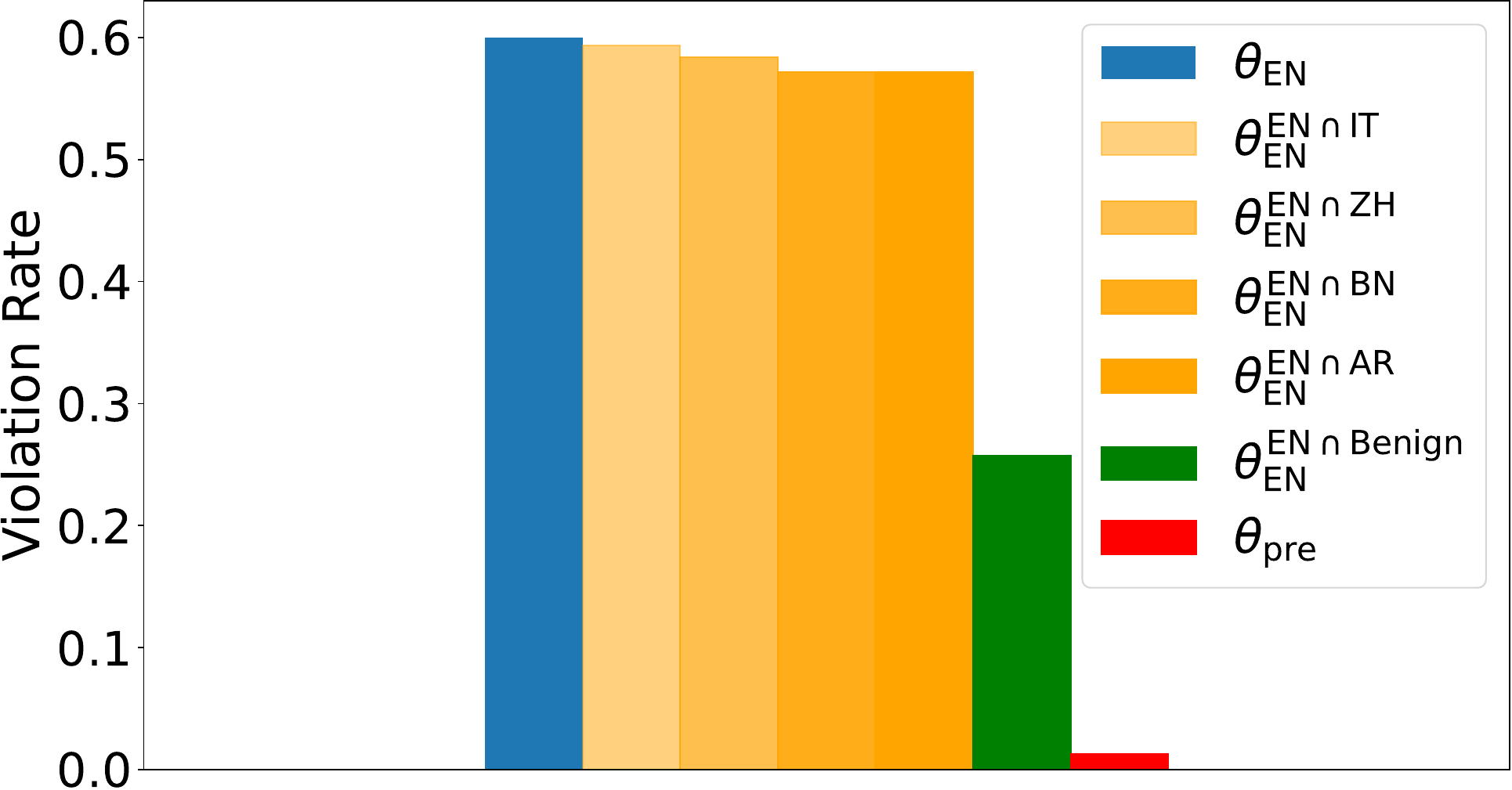}
    \caption{Llama-3.1-8B violation rates on the English language split of MultiJail after fine-tuning attack (blue) using English harmful data, stitching the bilingual intersection safety parameters localized by \ours (orange bars), benign datasets (green), and its original violation rate (red).
    }
    \vspace{-0.2cm}
    \label{fig:llama31-vr-hbr}
\end{figure}

\section{Details about freezing safety-related parameters experiments in \Cref{sec:freezing}}
\label{app:freezing}
In this lines we describe how we obtained the results we discussed in~\Cref{sec:freezing}.

Specifically, we start off by having a $\model{pre}$ and a $\modelnotext{l_{\text{ft}}}$, and we use \ours to localize an initial language-agnostic parameters $\binarymasknotext{\lpool}$. After this step, we freeze the parameters in $\model{pre}$ that correspond to the $1$s in $\binarymasknotext{\lpool}$ and perform the fine-tuning attack again, with the same configurations as described in~\Cref{app:attacks}, obtaining the new $\overline{\bm{\theta}}_{l_\text{ft}}$. Subsequently, we re-use \ours to localize the language-agnostic parameters $\overline{\bm{\gamma}}_{\lpool}$, in the attacked model $\overline{\bm{\theta}}_{l_\text{ft}}$, and maintain the same configurations mentioned in~\Cref{app:sil_details}. 

Now we verify the two properties discussed in~\Cref{sec:freezing}, and we first show in~\Cref{tab:global_overlappings} that $\binarymasknotext{\lpool} \cap \overline{\bm{\gamma}}_{\lpool} = 0$. Then we denote the \ours resulting stitched model to be 
$\bm{\theta}_{l_\text{ft}}^{\text{\ours}}$ and $\overline{\bm{\theta}}_{l_\text{ft}}^{\text{\ours}}$ before and after freezing respectively, and in~\Cref{tab:frozen-lang-agn-vr} we present the violation rate of $\overline{\bm{\theta}}_{l_\text{ft}}^{\text{\ours}}$. As it can be noticed, the new language-agnostic localized parameters retain the same level of violation capabilities, proving the alternative pathways hypothesis.

\end{document}